\documentclass[sigconf]{acmart}

\usepackage{booktabs} 
\usepackage{multirow}
\usepackage{flushend}
\usepackage[utf8]{inputenc}

\setcopyright{rightsretained}

\copyrightyear{2018}
\acmYear{2018} 
\setcopyright{iw3c2w3}
\acmConference[WWW '18 Companion]{The 2018 Web Conference Companion}{April 23--27, 2018}{Lyon, France}
\acmBooktitle{WWW '18 Companion: The 2018 Web Conference Companion, April 23--27, 2018, Lyon, France}
\acmPrice{}
\acmDOI{10.1145/3184558.3191823}
\acmISBN{978-1-4503-5640-4/18/04}
\editor{Jennifer B. Sartor}
\editor{Theo D'Hondt}
\editor{Wolfgang De Meuter}

\begin{document}
\title{Transfer Learning of Artist Group Factors \protect\\to Musical Genre Classification}

\author{Jaehun Kim}
\authornote{This is the corresponding author}
\orcid{0000-0001-5744-9034}
\affiliation{%
  \institution{Delft University of Technology}
  \streetaddress{6 Van Mourik Broekmanweg}
  \city{Delft}
  \state{} 
  \postcode{2600GA}
  \country{Netherlands}}
\email{j.h.kim@tudelft.nl} 

\author{Minz Won}
\authornote{This research was partially conducted during the author's internship at Kakao Corp.}
\orcid{}
\affiliation{%
  \institution{Universitat Pompeu Fabra}
  \streetaddress{}
  \city{Barcelona}
  \state{}
  \postcode{08005}
  \country{Spain}}
\email{minz.won@upf.edu}

\author{Xavier Serra}
\affiliation{%
  \institution{Universitat Pompeu Fabra}
  \streetaddress{}
  \city{Barcelona}
  \state{}
  \postcode{08005}
  \country{Spain}}
\email{xavier.serra@upf.edu}

\author{Cynthia C. S. Liem}
\affiliation{%
  \institution{Delft University of Technology}
  \city{Delft}
  \country{Netherlands}
}
\email{c.c.s.liem@tudelft.nl}


\renewcommand{\shortauthors}{J. Kim et al.}

\begin{abstract}
The automated recognition of music genres from audio information is a challenging problem, as genre labels are subjective and noisy. Artist labels are less subjective and less noisy, while certain artists may relate more strongly to certain genres. At the same time, at prediction time, it is not guaranteed that artist labels are available for a given audio segment. Therefore,  in this work, we propose to apply the transfer learning framework, learning artist-related information which will be used at inference time for genre classification. We consider different types of artist-related information, expressed through artist group factors, which will allow for more efficient learning and stronger robustness to potential label noise. Furthermore, we investigate how to achieve the highest validation accuracy on the given FMA dataset, by experimenting with various kinds of transfer methods, including single-task transfer, multi-task transfer and finally multi-task learning.
\end{abstract}

%
%

\ccsdesc[500]{Information systems~Music retrieval}
\ccsdesc[500]{Computing methodologies~Multi-task learning}
\ccsdesc[500]{Computing methodologies~Transfer learning}
\ccsdesc[500]{Computing methodologies~Neural networks}
%
%

\keywords{music information retrieval; multi-task learning;
transfer learning; neural network}

\maketitle

\section{Introduction}
\label{intro}

Learning to Recognize Musical Genre from Audio is a challenge track of \textit{The Web Conference 2018}. The main goal of the challenge is to predict musical genres of unknown audio segments correctly, by utilizing the FMA dataset ~\cite{benzi2016fma} as a training set. The challenge therefore focuses on a classification task.

In machine learning, many classification tasks, such as visual object recognition, consider objective and clearly separable classes. In contrast, music genres consider subjective, human-attributed labels. These may be inter-correlated (e.g.\ a \emph{rock} song may also be considered \emph{pop}, many \emph{classical} works are also \emph{instrumental}) and dependent of a user's context (e.g., a \emph{French rock} song is not \emph{International} to a French listener). Generally, no universal genre taxonomy exists, and even the definition of `genre' itself is problematic: what is usually understood as `genre' in Music Information Retrieval would rather be characterized as `style' in Musicology~\cite{liem12-dagstuhl}. This makes genre classification a challenging problem. In our work, considering the given labels in the challenge, we consider a musical genre to be a category that consists of songs sharing certain aspects of musical characteristics.

Commonly, music tracks are released with explicit mentioning of titles and artists. The identity of the artist does not suffer from semantic taxonomy problems, and can thus be considered as a more objective label than the genre label. At the same time, songs from the same artist tend to share prominent musical characteristics. Considering that an artist is commonly mapped into one or multiple specific genres, but not the whole universe of possible genres, and that the other way around, sets of artists can be seen as exemplars for certain music genres, the musical characteristics that identify an artist may also be key features of certain musical genres.

Therefore, it will be beneficial to exploit artist-related information in a genre classification task. At the same time, learning a direct mapping from artist identity  to genre label would not be practical. First of all, for an unknown audio segment for which a genre classification should be performed, the artist label may also not be available. Secondly, artist labels may not always be informative to a system, especially when an artist is newly introduced, so no previous history on the artist exists. Finally, an artist may have been active in multiple genres at once, but not be equally representative for all these genres. Given such constraints, we wish to employ  a learning framework which only requires artist labels at training time, but not at prediction time, and that will allow for the inclusion of newly introduced artists, for whom not much extra information is available beyond their songs.

In this work, we therefore present a multi-task transfer framework for using artist labels to improve a genre classification model. Assuming that artist labels are given for each track in the training set, these labels are used as side information, allowing a model to learn the mapping between audio and artists, while capturing patterns that might as well be useful for genre prediction.

It has been shown that music representations learned from raw artist labels can effectively transfer to other music-related tasks~\cite{park2017representation}. However, learning more than thousands of artists as individual classes is not efficient for at least two reasons:

\begin{itemize}
\item{Due to data sparsity, only a few tracks are assigned per class;}
\item{Despite the uniqueness of each artist, it can be beneficial to group them into clusters of similar artists, avoiding learning bottlenecks caused by large numbers of classes.}
\end{itemize}
To overcome these potential problems, we therefore apply a label pre-processing step, obtaining Artist Group Factors (AGF) as learning targets, rather than individual artist identities.

Finally, we train Deep Convolutional Neural Networks (DCNNs) employing different learning setups, ranging from targeting genre and various types of AGFs with individual networks, to employing a shared architecture as introduced in multiple previous Multi-Task Learning (MTL) works~\cite{caruana1998multitask, bengio2013representation, liu2015multi, bingel2017identifying, li2014heterogeneous, zhang2016deep, zhang2014facial,kim2018one}.

In the remainder of this paper, we first discuss an initial data exploration leading to our choice for AGFs (Section \ref{motivation}). Subsequently, we will give a detailed description of the proposed approach (Section \ref{method}), followed by a discussion of experimental settings (Section \ref{exp}). Finally, we will present our results (Section \ref{res}), followed by a short discussion and conclusion (Section \ref{disc}).


\section{Initial data exploration}
\label{motivation}
In the beginning of the challenge, we first explored the training data, and investigated a conventional data-driven approach using a DCNN for music genre classification, with genre labels as targets.

First of all, we had some concerns about the reliability of the genre annotations. As they were provided by users who uploaded the content, the users did not have access to a single genre taxonomy and unified annotation strategy. Thus, user-contributed annotations are expected to show more variability than annotations by experts. Furthermore, the dataset included 25,000 tracks from 5,152 unique albums. For 5,028 out of these 5,152 albums, genre annotations were made at the album level. While all tracks in an album can belong to a single genre, this is not always true. Indeed, we could discover examples of the case in which different tracks on the same album would belong to different genres, as well as multiple misannotations. Given these reliability issues, it is not guaranteed that by targeting these annotations only, generalized model performance for genre classification can be achieved.


To this end, while we will consider performance for direct (main top-)genre labels as targets (which we will denote as learning task category \texttt{g} in the remainder of this paper), in order to obtain more generalizable results obtained on more objective and consistent labeling data, we propose a multi-task transfer framework, introducing an Artist Group (AG) prediction task targeting AGFs.

\section{Methodology}
\label{method}

\subsection{Artist Group Factors}
\label{method:agf}

The main idea of extracting AGFs is to cluster artists based on meaningful feature sets that allow for aggregation at (and beyond) the artist level. For instance, one can collect genre labels from songs belonging to each artist, and then construct a Bag-of-Word (BoW) artist-level feature vector. Each dimension of the vector represents a genre, with the magnitude of the vector indicating genre frequency among a song collection. Alternatively, a BoW feature vector can be constructed by counting  latent `terms' belonging to each artist, which can be obtained by a dictionary learned from song-level or frame-level features through K-means clustering~\cite{lloyd1982least} or the Sparse Coding~\cite{coates2011importance} method.

\begin{figure*}
\includegraphics[width=0.8\textwidth]{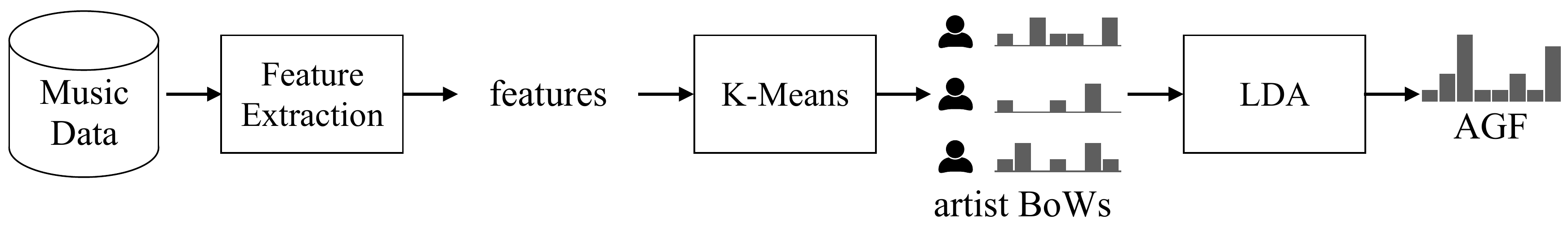}
\caption{Artist group factor extraction pipeline.}
\label{fig:agf}
\end{figure*}

Once artist-level BoW feature vectors are constructed, standard clustering methods such as K-Means, or more sophisticated topic modeling algorithms such as Latent Dirichlet Allocation (LDA)~\cite{blei2003latent} can be applied to find a small number of latent groups of artists: the AGFs for this particular feature set. This 2-step cascading pipeline is illustrated in Figure \ref{fig:agf}. 

In this work, we exploit four feature sets, which reflect different levels of musical and acoustical aspects of songs. From these feature sets, we obtain artist-level BoW vectors. Subsequently, LDA is applied to transform artist-level BoW vectors into dedicated AGF representations for the particular feature set. We will both consider these artist group prediction tasks and the main genre classification task within our learning framework: an overview summary is given in Table~\ref{tab:agf}.

\subsubsection{MFCCs}
\label{method:agf:mfcc}
Mel-Frequency Cepstral Coefficients (MFCCs), which are known to be efficient low-level descriptors for timbre analysis, were used as features of the artist grouping. The coefficients are initially calculated for short-time audio frames. Considering the coefficients over all audio frames of tracks for all artists, we build an universal dictionary of features using K-Means clustering. AGFs resulting from this feature set will belong to learning task category \texttt{m}.

\subsubsection{dMFCCs}
\label{method:agf:dmfcc}
Along with MFCCs, we also use time-deltas of MFCCs (first-order differences between subsequent frames), to consider the temporal dynamics of the timbre for the artist grouping. AGFs resulting from this feature set will be denoted by \texttt{d}.

\subsubsection{Essentia}
\label{method:agf:essentia}
We use song-level feature vectors from Essentia~\cite{bogdanov2013essentia}, which is a music feature extraction library. It extracts descriptors ranging from low-level features, such as statistics of spectral characteristics, to high-level features, including danceability~\cite{herrera2005detrended} or semantic features learned from the data. After filtering descriptor entries that include missing values or errors, we obtained a 4374-dimensional feature vector per track. Before training a dictionary, we apply quantile normalization: a rank-based normalization process that transforms the distribution of the given features to follow a target distribution~\cite{doi:10.1198/016214501753381814}, which we set to be a normal distribution in this case. AGFs resulting from this feature set will belong to learning task category \texttt{e}.

\subsubsection{Subgenres}
\label{method:agf:subg}
%
%
We also use the 150 genre labels, including sub-genres, as a pre-defined dictionary for semantic description. For these, we directly build artist-level BoW vectors by aggregating all the genre labels from tracks by an artist. AGFs resulting from this feature set will belong to learning task category \texttt{s}.

\begin{table}[]
\centering
\caption{Details of Learning Targets}
\label{tab:agf}
\begin{tabular}{lcccc}
id & Category & Source    & Dictionary               & Dimension         \\ \hline\hline
\texttt{g} & Main & Genre    & N / A				& 16 \\ \hline
\texttt{m} & \multirow{4}{*}{AGF} & MFCC     & \multirow{3}{*}{K-means} & 25 \\
\texttt{d} &  & dMFCC    &                          & 25                     \\
\texttt{e} &  & Essentia~\cite{bogdanov2013essentia} &     & 4374                     \\
\texttt{s} &  & Subgenre    & N / A                    & 150                 \\ \hline
\end{tabular}
\end{table}

\begin{table}[]
\centering
\caption{Network Architectures for Encoder $f$}
\label{tab:dcnn}
\begin{tabular}{ll}
Layers & Output shape                             \\
\hline\hline
Input layer               & $128\times43\times1$  \\ 
\hline
Conv $5\times5$, ELU      & $128\times43\times16$ \\ 
MaxPooling $2\times1$     & $64\times43\times16$  \\
\hline
Conv $3\times3$, BN, ELU  & $64\times43\times32$  \\ 
MaxPooling $2\times2$     & $32\times21\times32$  \\
Dropout (0.1)             & $32\times21\times32$  \\
\hline
Conv $3\times3$, ELU      & $32\times21\times64$  \\ 
MaxPooling $2\times2$     & $16\times10\times64$  \\ 
\hline
Conv $3\times3$, BN, ELU  & $16\times10\times64$  \\
MaxPooling $2\times2$     & $8\times5\times64$    \\
Dropout (0.1)             & $8\times5\times64$    \\
\hline
Conv $3\times3$, ELU      & $8\times5\times128$   \\ 
MaxPooling $2\times2$     & $4\times2\times128$   \\
\hline
Conv $3\times3$, ELU      & $4\times2\times256$   \\ 
\hline
Conv $1\times1$, BN, ELU  & $4\times2\times256$   \\
\hline
GlobalAveragePooling, BN  & 256                   \\
\hline
Dense, BN, ELU            & 256                   \\
Dropout (0.5)             & 256                   \\
\hline
Output layer 16 or 40     & 16 or 40              \\
\end{tabular}
\end{table}

\subsection{Network Architectures}
\label{method:arch}
\begin{figure}
	\centering
		\includegraphics[height=0.3\textheight]{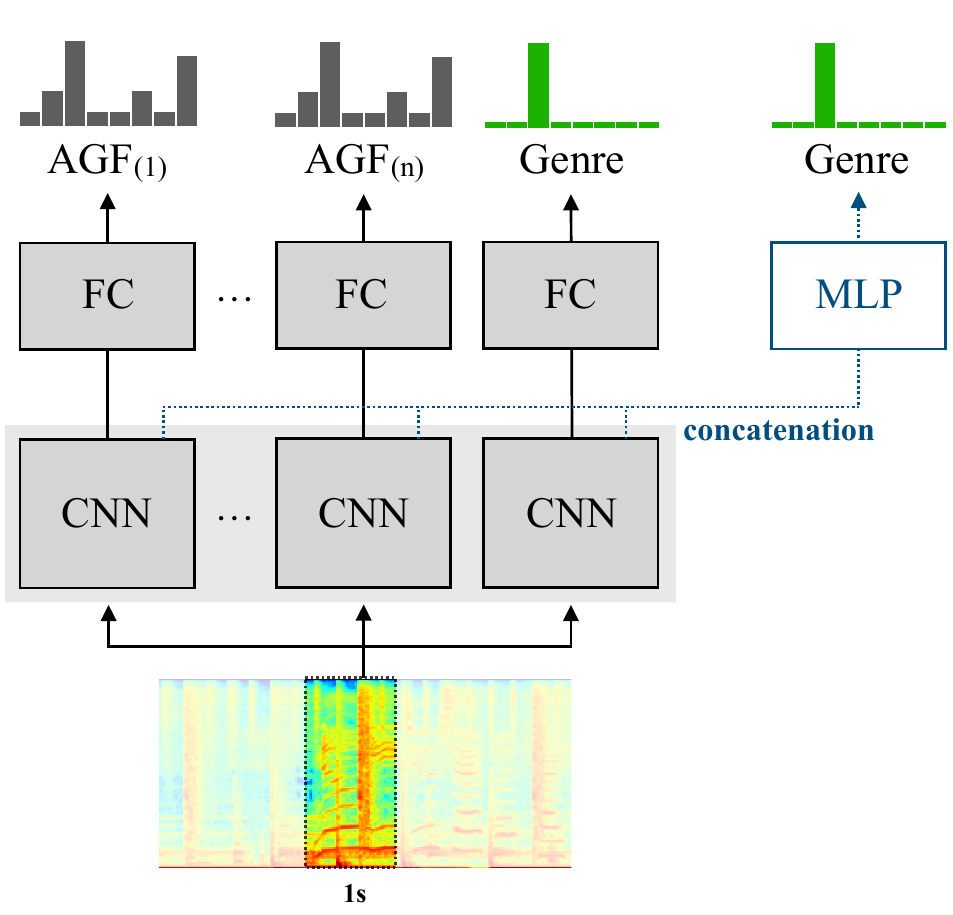}
	\caption{Illustration for the transfer learning scenario. Dotted lines indicate the setup for the multilayer perceptron for performing final genre classification.}
	\label{fig:transfer}
\end{figure}

The architecture of the proposed system can be divided into two parts, as shown in Figure \ref{fig:transfer}. We first train multiple DCNNs, targeting the various categories of learning targets (genres or various AGFs). Subsequently,  transfer takes place: a multilayer perceptron (MLP) for the final genre classification is trained, utilizing features that were derived from the previously trained DCNNs.

\subsubsection{DCNN}
\label{method:DCNN}
We adapted DCNN models to obtain transferable features for  genre classification (Table \ref{tab:dcnn}). The input size of the input layer is $128\times$43, which is the size of a spectrogram with 128 mel bins and 43 samples (1 second of audio). After the input layer, there are seven convolutional layers followed by a max-pooling layer, except for the last two layers. The first convolutional layer has $5\times5$ kernels and the last convolutional layer has $1\times$1 kernels. Except for those two layers, all convolutional layers have $3\times$3 kernels. Outputs of the last convolutional layer are subsampled by global-average-pooling. Finally, they are connected to two dense layers for predicting AGF clusters or genres. Batch normalization~\cite{ioffe2015batch} and dropouts~\cite{srivastava2014dropout} are sparsely used to prevent overfitting. Exponential Linear Unit (ELU)~\cite{clevert2015fast} is used as an activation function for the convolutional layers and Softmax is used for the output layer.

\subsubsection{Shared Architecture}
\label{method:shared_arch}

Considering that lower layers of DCNNs usually capture lower-level features such as edges from images or spectrograms, we hypothesized that sharing lower layers among the various DCNNs can be effective under the scenario where multiple learning sources are available. With this approach, one can expect that it not only ensures sufficient specialization on task-specific upper layers, but also  benefits from regularization effects on lower layers\cite{kim2018one}. Joint learning of multiple tasks with shared layers can prevent the shared layer to overfit for a specific task, instead learning underlying factors that have commonalities required across tasks~\cite{caruana1998multitask,liu2015representation}.

Throughout the experiment, we used the shared architecture that shares only the first convolutional block. It consists of the first convolutional and the max-pooling layer. For brevity, for the remainder of the paper, we use Single-Task Nets (STNs) and an Multi-Task Net (MTN) to refer to the non-shared networks and shared networks respectively.

\subsubsection{Transfer method}
\label{method:transfer}
The proposed system learns and predicts a genre of an input spectrogram by transferring pre-trained features from Section~\ref{method:DCNN}. We trained an MLP with a single hidden layer; the size of the hidden layer was 1024. ELU non-linearity was used for the hidden layer and Softmax was used for the output layer. Dropouts of 50\% were applied for the input layer and a hidden layer.

Note that for both the feature learning phase and the transfer learning phase, we keep using a segment-wise learning approach. Only at the final inference step, we aggregate all the segment-level predictions, by taking the average of each segment's predicted probability for the genres.

\subsubsection{Training}
\label{method:train}
At training time, we iteratively update the model parameters with the mini-batch stochastic gradient descent method using the Adam algorithm~\cite{kingma2014adam}. For data augmentation, we randomly crop 1-second excerpts from the entire track included in the mini-batch. We use 64 samples per batch and set the learning rate to 0.001 across the experiments.

For comparison between methods, experiments are run with a fixed number of epochs. We set 1000 epochs for an MTN and 200 for STNs. Since we took a similar stochastic update algorithm to~\cite{liu2015multi} for the shared architecture, for the number of updates for task-specific layers in a shared network, the number of epochs used for training non-shared networks should be multiplied with the number of involved learning tasks. For the transfer learning phase, we also set the number of epochs to train the MLP to 50.

\subsection{Pre-processing}
\label{method:preproc}
We use mel spectrograms as the input representation for the neural networks. We extract 128-dimensional mel spectra for audio frames of 46ms, with 50\% overlap with adjacent frames. To enhance lower-intensity levels of input mel spectrograms at higher frequencies, we take dB-scale log amplitudes of each mel spectrum.

\subsection{Implementation Details}
\label{method:imple}

The experiments were run on GPU-accelerated hardware and software environments. We used Lasagne~\cite{sander_dieleman_2015_27878}, Theano~\cite{2016arXiv160502688short} and Keras~\cite{chollet2015} as main experimental frameworks\footnote{The main code for the experiment can be found in \url{https://github.com/eldrin/Lasagne-MultiTaskLearning}}. We used a number of different GPUs, including NVIDIA GRID-K2, NVIDIA GTX 1070, NVIDIA TITAN X.

\section{Experiments}
\label{exp}
To investigate the effectiveness of various types of AGFs for transfer learning, we trained all 31 possible combinations of given learning tasks, including AGFs (\texttt{m}, \texttt{d}, \texttt{e}, \texttt{s}) and main top-genre labels (\texttt{g}). For each run, to investigate the optimal feature architecture, we tested both shared networks and separate networks for each learning task. This leads to a total number of 62 cases, including all the combinations of learning tasks per network architecture.

However, in all cases in which multiple tasks are considered, the networks  have a larger number of parameters compared to the case in which a network focuses on a single task. With a subsequent experiment, we therefore tried to verify the effect of more parameters and larger networks vs. \ the effect of using more tasks. To this end, we train wide Single Task Networks (wSTNs), targeting only genre, but having an equal number of parameters to the MTNs/STNs targeting multiple tasks. Finally, with respect to the number of tasks involved, we compare the best performance of MTNs/STNs to the performance of wSTNs with the same number of parameters.

As for the AGFs using song-level or frame-level features, we trained K-means algorithms employing 2048 clusters. We observed that lower numbers of clusters (e.g.\ 1024) can cause artists with few tracks to get a zero vector as artist-level BoW representation, due to data sparsity. Throughout the experiments, we used a fixed number of latent artist groups, set to 40.

Finally, for the internal evaluation, we divided the given training dataset employing a stratified random 85/15 split.

\section{Results}
\label{res}

\subsection{Multiple Learning Tasks in STN vs.\ MTN}
\label{res:stl}

\begin{figure}
	\centering
		\includegraphics[width=0.5\textwidth]{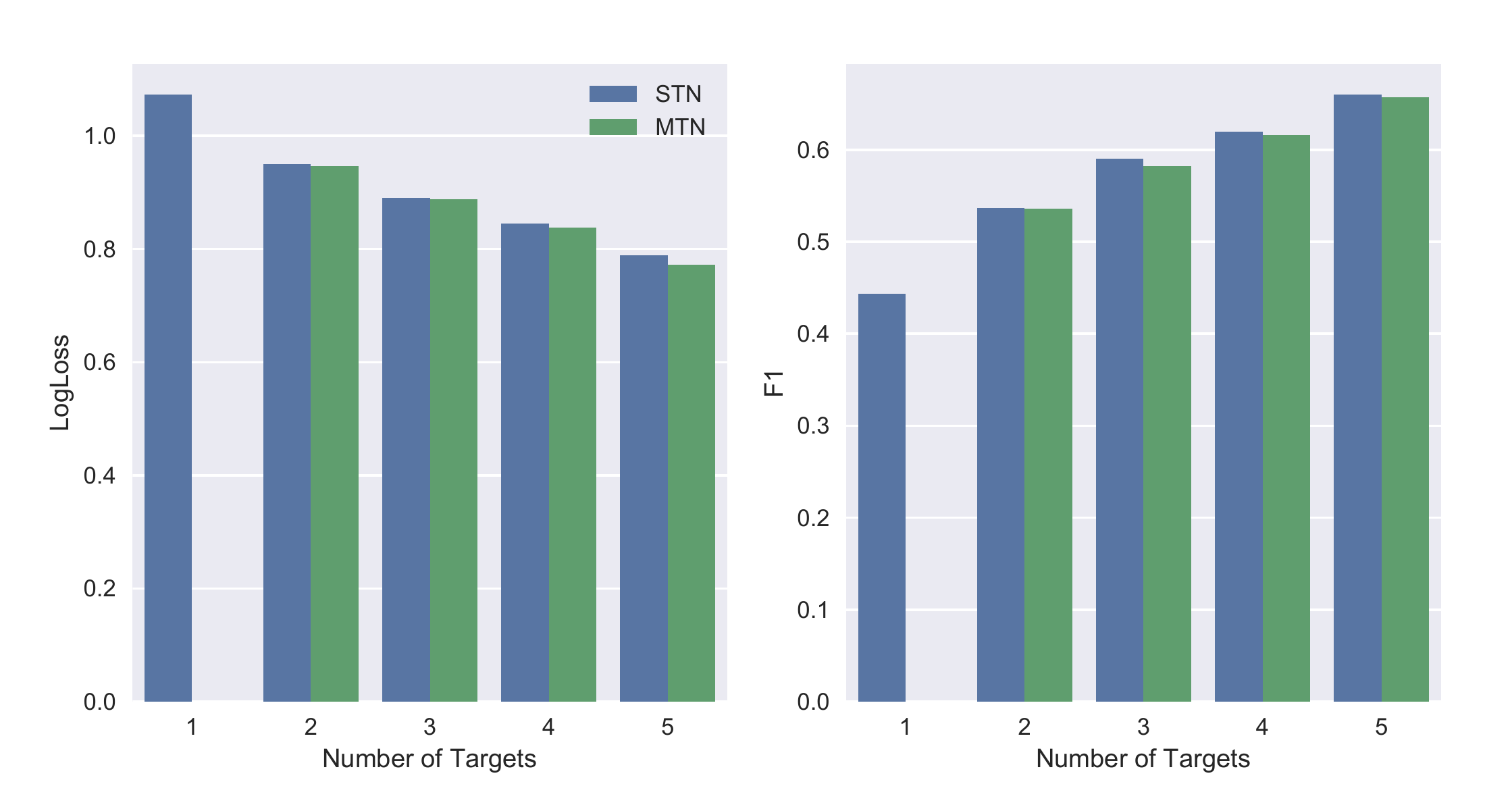}
	\caption{Average performance for the number of tasks involved in feature learning}
	\label{fig:effect_of_n}
\end{figure}

\begin{table}[]
\centering
\caption{Comparison of the average performance with or without the main task}
\label{tab:is_main}
\begin{tabular}{lllll}
             & \multicolumn{2}{c}{LogLoss}                       & \multicolumn{2}{c}{F1}                            \\ \cline{2-5} 
 & \multicolumn{1}{c}{STN} & \multicolumn{1}{c}{MTN} & \multicolumn{1}{c}{STN} & \multicolumn{1}{c}{MTN} \\ \hline\hline
without \texttt{g}  & 1.0079                  & 0.9618                  & 0.4932                  & 0.5168                  \\
with \texttt{g}     & \textbf{0.8540}         & \textbf{0.8486}         & \textbf{0.6154}         & \textbf{0.6155}         \\ \hline
\end{tabular}
\end{table}

In general, we observe that the number of learning tasks has a positive effect on both performance metrics. As shown in Table~\ref{tab:is_main}, it also is found that cases in which the main top-genre classification are included yield better results in comparison to other combinations of tasks.

Considering STN vs.\ MTN, on the log loss metric, MTN shows better results, but in the case of the f1-measure, the opposite is shown. Generally, considering the number of learning tasks and absolute magnitude of differences, the difference observed between the two methods cannot be deemed significant; more experiments with additional datasets and multiple splits would be needed to assess whether statistically significant differences between STN vs.\ MTN approaches can be obtained.

For both STN and MTN, the best performance we achieved uses all the learning tasks, as shown in the last row of Table \ref{tab:main_res}.

\begin{table}[]
\centering
\caption{The performance of various combinations of AGFs and the top-level main genre target as a feature learning task.}
\label{tab:main_res}
\begin{tabular}{lllll}
      & \multicolumn{2}{c}{STN}           & \multicolumn{2}{c}{MTN}           \\ \cline{2-5} 
      & LogLoss         & F1              & LogLoss         & F1              \\ \hline\hline
\texttt{g}     & 0.8891          & 0.5963          & \multirow{5}{*}{N/A}  &  \multirow{5}{*}{N/A}         \\
\texttt{m}     & 1.1812          & 0.3581          &           &           \\
\texttt{d}     & 1.0987          & 0.3967          &           &           \\
\texttt{e}     & 1.2542          & 0.3437          &           &           \\
\texttt{s }    & 0.9404          & 0.5218          &           &           \\
\texttt{gs}    & 0.8606          & 0.6114          & 0.8578          & 0.6190          \\
\texttt{ge}    & 0.8811          & 0.5953          & 0.8792          & 0.5996          \\
\texttt{gd}    & 0.8845          & 0.5898          & 0.8803          & 0.5955          \\
\texttt{gm}    & 0.8874          & 0.5957          & 0.8813          & 0.6037          \\
\texttt{se}    & 0.9124          & 0.5537          & 0.9079          & 0.5502          \\
\texttt{sd}    & 0.9191          & 0.5601          & 0.9146          & 0.5412          \\
\texttt{sm}    & 0.9260          & 0.5581          & 0.9283          & 0.5458          \\
\texttt{ed }   & 1.0557          & 0.4433          & 1.0422          & 0.4399          \\
\texttt{em}    & 1.1186          & 0.4244          & 1.1060          & 0.4376          \\
\texttt{dm}    & 1.0583          & 0.4373          & 1.0704          & 0.4280          \\
\texttt{gse}   & 0.8361          & 0.6255          & 0.8335          & 0.6277          \\
\texttt{gsd}   & 0.8579          & 0.6280          & 0.8519          & 0.6150          \\
\texttt{gsm}   & 0.8486          & 0.6289          & 0.8541          & 0.6153          \\
\texttt{ged}   & 0.8528          & 0.6051          & 0.8601          & 0.6067          \\
\texttt{gem}   & 0.8645          & 0.5988          & 0.8701          & 0.6056          \\
\texttt{gdm}   & 0.8773          & 0.5985          & 0.8845          & 0.5941          \\
\texttt{sed}   & 0.8965          & 0.5818          & 0.8867          & 0.5640          \\
\texttt{sem}   & 0.9104          & 0.5834          & 0.8889          & 0.5668          \\
\texttt{sdm}   & 0.9211          & 0.5629          & 0.9109          & 0.5572          \\
\texttt{edm}   & 1.0359          & 0.4879          & 1.0365          & 0.4675          \\
\texttt{gsed}  & 0.8211          & 0.6343          & 0.8132          & 0.6328          \\
\texttt{gsem}  & 0.8264          & 0.6352          & 0.8172          & 0.6284          \\
\texttt{gsdm}  & 0.8407          & 0.6379          & 0.8288          & 0.6170          \\
\texttt{gedm}  & 0.8466          & 0.6053          & 0.8450          & 0.6152          \\
\texttt{sedm } & 0.8906          & 0.5856          & 0.8875          & 0.5870          \\
\texttt{gsedm} & \textbf{0.7894} & \textbf{0.6599} & \textbf{0.7727} & \textbf{0.6571} \\ \hline
\end{tabular}
\end{table}

\subsection{Networks for Multiple Learning Tasks vs.\ Large Network on a Single Task}
\label{res:fat_vs_transfer}
We also compared the performance between the best STNs and MTNs for a given number of learning tasks, versus the performance of a wSTN that has equal model capability to these multi-task setups in terms of parameters and architecture, but only is trained on direct main top-genre classification. The corresponding results are shown in Table \ref{tab:multi_single}. It can be seen that MTN representations yield better performance on the log loss metric when all 5 learning tasks (all AGFs and the main top-genre) are used, although at the same time, wSTN performs better when considering the f1-measure for the case in which 2 learning tasks are used. In other cases, differences between the setups appear marginal; further experiments would be needed to assess  whether STNs/MTNs will give significant performance boosts in case a larger set of tasks would be considered.

\begin{table}[]
\centering
\caption{Comparison between wSTN (single genre classification task) and STN/MTN setups (multiple tasks) learning setups. The reported performances of STN and MTN consider the task combinations for which the best performance was obtained, given the mentioned number $N$ of tasks.}
\label{tab:multi_single}
\begin{tabular}{lllllll}
  & \multicolumn{3}{c}{LogLoss}       & \multicolumn{3}{c}{F1}            \\ \cline{2-7} 
N & wSTN   & STN    & MTN             & wSTN            & STN    & MTN    \\ \hline\hline
2 & 0.8688 & 0.8606 & 0.8578          & 0.6071          & 0.6114 & 0.6190 \\
3 & 0.8546 & 0.8361 & 0.8335          & \textbf{0.6629} & 0.6289 & 0.6277 \\
4 & 0.8278 & 0.8211 & 0.8132          & 0.6451          & 0.6352 & 0.6328 \\
5 & 0.8290 & 0.7893 & \textbf{0.7727} & 0.6528          & 0.6599 & 0.6571 \\ \hline
\end{tabular}
\end{table}

\section{Discussion \& Conclusion}
\label{disc}
In this work, we proposed including several categories of low-rank AGFs, expressing artist-level information, into the task of classifying music genre based on musical audio. Our experimental results support the hypothesis that by targeting different categories of AGFs, deep networks can learn features from musical audio that can meaningfully support genre classification. The inclusion of multiple parallel learning tasks considering different AGF categories, and the inclusion of both genre- and AGF-based tasks in a multi-task setup, also both seem beneficial, although further work will need to be done to assess whether observed effects are truly significant. For this, other datasets will have to be included for training and testing; furthermore, alternative cluster algorithms and clustering parameters should be investigated to achieve the most robust AGF-based features.



\begin{acks}
This work was carried out on the Dutch national e-infrastructure with the support of
SURF Cooperative. And this work is partially supported by the Maria de Maeztu Programme (MDM-2015-0502). We further acknowledge the computing support of Kakao Corporation.
\end{acks}

\bibliographystyle{ACM-Reference-Format}
\bibliography{sample-bibliography}

\end{document}